\documentclass[runningheads]{llncs}

\usepackage{eccv}

\usepackage{eccvabbrv}

\usepackage{graphicx}
\usepackage{booktabs}

\usepackage[accsupp]{axessibility}  

\usepackage{hyperref}

\usepackage{orcidlink}
\usepackage{multirow}
\usepackage{multicol}
\usepackage{wrapfig}

\usepackage{algorithm}
\usepackage{algorithmic}
\usepackage{amsmath}
\usepackage{amssymb}
\DeclareMathOperator*{\argmax}{arg\,max}

\usepackage[table]{xcolor}

\begin{document}

\title{HiMAC: Hierarchical Macro–Micro Learning for Long-Horizon LLM Agents}

\titlerunning{Abbreviated paper title}

\setcounter{footnote}{1}
\author{Hongbo Jin\inst{\dagger}
\and
Rongpeng Zhu\inst{\dagger}
\and
Jiayu Ding
\and
Guibo Luo
\and
Ge Li\inst{\ast}
}

\authorrunning{F.~Author et al.}

\institute{School of Electronic and Computer Engineering\\ Peking University}

\maketitle
\footnotetext{
\scriptsize 
{$\dagger$}: Equal key contributions.\\
{$\ast$}: Corresponding author. 
}

\begin{abstract}
Large language model (LLM) agents have recently demonstrated strong capabilities in interactive decision-making, yet they remain fundamentally limited in long-horizon tasks that require structured planning and reliable execution. Existing approaches predominantly rely on flat autoregressive policies, where high-level reasoning and low-level actions are generated within a single token sequence, leading to inefficient exploration and severe error propagation over extended trajectories.
In this work, we propose \textbf{HiMAC}, a hierarchical agentic RL framework that explicitly decomposes long-horizon decision-making into macro-level planning and micro-level execution.
HiMAC models reasoning as a structured blueprint generation process followed by goal-conditioned action execution,
enabling robust long-horizon planning within LLM-based agents.
To train this hierarchy efficiently, we introduce a critic-free \textbf{hierarchical policy optimization} paradigm that extends group-based reinforcement learning to bi-level structures through hierarchical relative advantage estimation. Furthermore, we propose an \textbf{iterative co-evolution} training strategy that alternates between planner exploration and executor adaptation, mitigating the non-stationarity inherent in hierarchical learning.
Extensive experiments on ALFWorld, WebShop, and Sokoban demonstrate that HiMAC consistently outperforms strong prompting and reinforcement learning baselines, achieving state-of-the-art performance and substantially improved sample efficiency across both text-based and visually grounded environments. Our results show that introducing structured hierarchy, rather than increasing model scale alone, is a key factor for enabling robust long-horizon agentic intelligence.
  \keywords{LLM Agent \and Hierarchical RL \and Long-Horizon Planning}
\end{abstract}

\section{Introduction}
\label{sec:intro}
The rapid evolution of Large Language Models (LLMs) has sparked a paradigm shift in agentic AI, transitioning from passive answering systems to autonomous agents capable of active environment interaction \cite{wang2024survey, sapkota2025ai, sang2025beyond}.
By grounding linguistic reasoning in sequential decision-making, LLM-based agents have demonstrated impressive proficiency in short-horizon tasks\cite{li2022pre, chen2024efficient}.
However, current LLM agents exhibit three coupled failure modes in long-horizon settings: exponential exploration complexity, delayed credit assignment, and semantic drift across extended reasoning trajectories.

The prevailing landscape of agentic control is dominated by "flat" policy architectures\cite{huang2024understanding, chen2024can, erdogan2025plan}.
In these paradigms, a single autoregressive model is tasked with generating both high-level thoughts and low-level actions in a dense, token-by-token manner.
While straightforward, this monolithic formulation suffers fundamentally from the curse of dimensionality in exploration~\cite{yao2023tree, kulkarni2016hierarchical, xie2024travelplanner, chang2024agentboard}.
Under this setting, the agent must navigate a vast combinatorial search space using myopic next-token prediction.
In such flat trajectories, errors do not merely occur but also propagate exponentially.
A minor syntactic deviation in an early step often cascades into irreversible failure states, causing the agent to lose track of the global goal.
This limitation indicates that relying solely on the inherent reasoning capabilities of generic LLMs is insufficient; structural inductive biases are required to decouple global planning from local control \cite{dayan1992feudal, parr1997reinforcement, sutton1999between, barto2003recent}.
\begin{figure}[h]
    \centering
    \includegraphics[width=\linewidth]{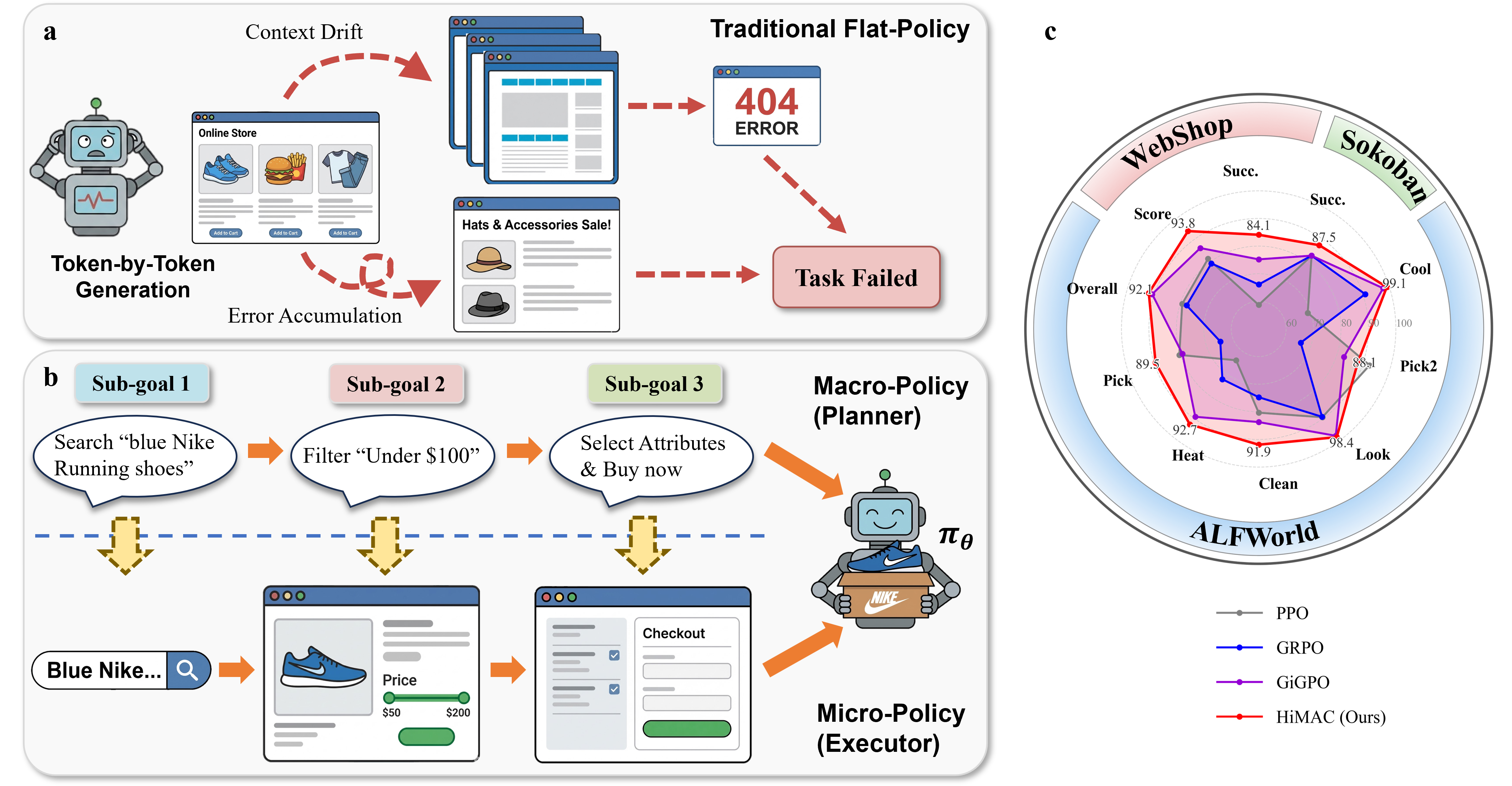}
    \caption{A typical case comparsion in WebShop environment and overall performance. (a) Flat-policy architectures easily get lost in irrelevant observations (Context Drift). (b) HiMAC addresses this exploration inefficiency through a bi-level decoupled architecture (Planner and Executor sharing the same parameters $\pi_{\theta}$), mapping structured blueprints to executable atomic actions. (c) HiMAC yields consistent state-of-the-art performance against advanced RL reasoning baselines across multiple environments.}
    \label{fig:case compare}
\end{figure}

This calls for a fundamental shift in how LLM agents are structured.
We propose \textbf{HiMAC} (\textbf{Hi}erarchical \textbf{M}acro-Micro \textbf{A}gentic \textbf{C}ontrol),
a framework that explicitly organizes the agent as a two-level cooperation.
The Macro-Policy operates as a strategic planner: given a task instruction,
it searches over a latent semantic space to produce a blueprint—a structured sequence of natural language sub-goals that decompose the long-horizon objective into tractable milestones.
Conditioned on this blueprint,
the Micro-Policy operates as a focused executor, generating atomic actions for each sub-goal in sequence.
As illustrated in Fig. \ref{fig:case compare},
this separation fundamentally changes the nature of the exploration problem:
rather than navigating an exponentially large joint action-reasoning space in a single flat trajectory. 

Realizing this hierarchical vision, however, introduces a formidable optimization challenge. Training the two levels jointly constitutes a non-stationary bi-level problem: the Macro-Policy must learn to propose sub-goals for an executor whose capabilities are simultaneously evolving, while the Micro-Policy must learn to execute plans whose quality is shifting as the planner improves.
Standard RL approaches attempt to resolve such instability via auxiliary Value Networks (Critics) \cite{schulman2017proximal, rafailov2023direct, raileanu2021decoupling,sutton1998reinforcement}, but learning accurate value functions in the high-dimensional, sparse semantic space of language is notoriously sample-inefficient and prone to divergence.
Without stable optimization, hierarchical decomposition offers no practical benefit over flat policies.

Driven by these challenges, HiMAC introduces two tightly coupled technical innovations. (i) \textbf{Critic-Free Hierarchical Policy Optimization}: We extend Group Relative Policy Optimization (GRPO) \cite{shao2024deepseekmath} to a bi-level structure by constructing level-specific comparison groups—sampled blueprints are evaluated against peer blueprints, and execution trajectories are evaluated against peer trajectories conditioned on the same blueprint. This hierarchical grouping yields low-variance advantage estimates tailored to each level, enabling precise credit assignment without any parametric value network.
(ii) \textbf{Iterative Co-Evolution Training}: To resolve non-stationarity, we decouple the two optimization objectives into alternating phases. In the Macro-Exploration Phase, the planner is updated while the executor runs in inference mode, providing a deterministic reward signal that drives the planner toward practically achievable blueprints. In the Micro-Adaptation Phase, a high-confidence blueprint is fixed as a constant condition, and the executor is updated in isolation. This alternation converts the unstable bi-level problem into a sequence of stationary single-level updates, naturally inducing a curriculum where the planner progressively proposes more complex strategies as executor proficiency grows.


We evaluate HiMAC across three benchmarks specifically chosen to stress-test different facets of hierarchical agentic control: ALFWorld \cite{shridhar2020alfworld} (multi-step embodied reasoning in household environments), WebShop \cite{yao2022webshop} (long-horizon web navigation with noisy, high-dimensional observations), and Sokoban \cite{SchraderSokoban2018} (visually-grounded spatial planning requiring precise sequential reasoning).
Against strong prompting techniques \cite{yao2022react,shinn2023reflexion} and recent critic-free RL methods \cite{shao2024deepseekmath, feng2025group} including GRPO and GiGPO,
HiMAC achieves state-of-the-art results on all three—most notably a 16\% gain over the strongest RL baseline on WebShop (83.4\% vs. 67.4\%),
a benchmark where flat policies are most severely penalized by context drift.
Beyond raw performance, HiMAC reaches comparable success thresholds with substantially fewer training iterations, and qualitative analysis reveals that the Macro-Policy spontaneously develops self-verification behaviors absent from any flat baseline.
Together, these results support a broader conclusion: structured hierarchy provides complementary and substantial gains beyond those achievable through model scale alone.


In summary, our main contributions are as follows:
\begin{itemize}
    \item We propose \textbf{HiMAC}, a hierarchical framework that decomposes long-horizon agentic tasks into Macro-level blueprint generation and Micro-level goal-conditioned execution, fundamentally reducing the exploration complexity and error propagation of flat policy architectures.
    \item We introduce a Critic-Free \textbf{Hierarchical Policy Optimization} objective and an \textbf{Iterative Co-Evolution} training strategy, extending group-based RL to a bi-level structure for precise level-specific credit assignment, while alternating Macro-Exploration and Micro-Adaptation phases to stabilize the non-stationary bi-level optimization.
    \item  Experiments on various challenging benchmarks demonstrate state-of-the-art performance across text-based and visually-grounded environments, with up to 16\% gain over the strongest RL baseline and superior sample efficiency.
\end{itemize}

\section{Related Work}
\label{sec:relat}


\paragraph{LLM Agents.}
Large language models (LLMs) \cite{achiam2023gpt, zhao2023survey} and, increasingly, Vision-Language Models (VLMs) \cite{liu2023visual, team2023gemini} have rapidly evolved into versatile autonomous agents capable of operating in open-ended text and visually-grounded environments \cite{wang2023voyager, zitkovich2023rt, zhang2024codeagent,jin2026tir}.
Early prompting methods, such as ReAct \cite{yao2022react} and Reflexion \cite{shinn2023reflexion}, proved effective for short-horizon tasks by interleaving reasoning with actions.
However, these "flat" autoregressive approaches struggle severely in long-horizon scenarios.
In these settings, myopic next-token prediction often causes minor deviations to cascade into irreversible failures.
Because these errors fundamentally stem from entangling high-level reasoning with low-level execution in a single token stream, relying solely on frozen foundation models is insufficient.
To robustly decouple global semantic planning from local control, agents require explicitly hierarchical architectures combined with corresponding policy optimization strategy \cite{ouyang2022training, liu2023pre,goyal2022inductive,battaglia2018relational}.


\paragraph{Agentic RL.}
Applying RL to LLM-based agents has emerged as a promising direction for sequential decision-making \cite{rawles2023androidinthewild, brockman2016openai, zhai2024fine, tan2024true, wen2024reinforcing, bai2024digirl,jin2025videocurl,jin2025videomem}.
Traditional actor-critic methods such as PPO \cite{schulman2017proximal} have been adopted for embodied and web navigation agents \cite{shridhar2020alfworld, yao2022webshop}, but learning accurate value functions over high-dimensional language states with sparse rewards is notoriously sample-inefficient and prone to divergence. This has motivated a shift toward critic-free, group-based objectives such as RLOO \cite{ahmadian2024back} and GRPO \cite{shao2024deepseekmath}, which estimate relative advantages within sampled groups and yield more stable convergence. GiGPO \cite{feng2025group} and RAGEN \cite{wang2025ragen} further extend this paradigm to multi-turn agentic settings. However, all of these methods optimize a monolithic policy over the joint space of reasoning and action tokens, conflating high-level semantic intent with low-level execution in a single flat trajectory. As a result, they inherit the exponential exploration complexity of long-horizon tasks and remain susceptible to error propagation across extended interaction sequences—limitations that hierarchical decomposition is specifically designed to address.


\paragraph{Hierarchical Reinforcement Learning.}
Classical HRL \cite{nachum2018data, li2025hiplan, ahn2022can, zhu2023ghost, sutton1998reinforcement, dayan1992feudal} addresses long-horizon tasks via temporal abstraction. Recent robotics approaches further mitigate hierarchical non-stationarity via preference optimization \cite{singh2024dipper} or alternating training \cite{lu2025reinforcement}, yet they remain grounded in compact, predefined state spaces distinct from the open-ended reasoning required by LLMs.
In the realm of LLM agents, works largely rely on prompt engineering without parameter optimization (e.g., HiAgent \cite{hu2025hiagent}), or depend heavily on offline datasets and fixed primitives (e.g., LARAP \cite{zhang2025llms}, GLIDER \cite{gao2025solving}). Similarly, while advanced RL methods target long-horizon tasks, they often optimize flat policies (e.g., AgentGym-RL \cite{xi2025agentgym}) or rely on offline preference learning (e.g., HPL \cite{hu2025divide}). CoDA \cite{liu2025coda} introduces a context-decoupled framework for multi-hop QA via joint optimization. In contrast, HiMAC bridges existing gaps in embodied tasks by decoupling decision-making into a Macro-Planner and Micro-Executor, directly optimizing natural-language parameters through a stable, critic-free hierarchical objective driven by iterative co-evolution.

\section{Methodology}
\label{sec:method}

\subsection{Problem Formulation}


We formulate the long-horizon agentic task as a Goal-Conditioned POMDP \cite{kaelbling1998planning, schaul2015universal}, where an agent receives partial observations $o_t \in \mathcal{O}$ and executes atomic actions $a_t \in \mathcal{A}$ to maximize the expected return given a natural language instruction $x$.

Standard approaches model the policy $\pi(a_t | o_{\le t}, x)$ as a flat autoregressive process. However, for long-horizon tasks, the probability of optimal trajectory generation $P(\tau|x)$ vanishes exponentially with horizon length $T$. To address this, we introduce a Structured Blueprint $\mathbf{z}$ as a discrete semantic variable, representing a complete high-level plan decomposed from $x$.
We formulate the trajectory generation as a Hierarchical Process where the joint distribution of the blueprint $\mathbf{z}$ and atomic actions $\mathbf{a}$ is factorized as:
 \begin{equation}
  P(\mathbf{a}, \mathbf{z} | x) = \underbrace{\pi_{\text{macro}}(\mathbf{z} | x)}_{\text{Planning}} \cdot \prod_{t=1}^{T} \underbrace{\pi_{\text{micro}}(a_t | o_{\le t}, \mathbf{z}_{\phi(t)})}_{\text{Execution}}  
\end{equation}
where $\mathbf{z} \in \mathcal{Z}$ denotes a sequence of natural language sub-goals $\{g_1, \dots, g_K\}$,
and $\phi(t)$ maps the timestep to the active sub-goal index.

This formulation decouples the problem into two optimization objectives:
(i) Macro-Optimization: Finding the optimal blueprint $\mathbf{z}^* = \arg\max_{\mathbf{z}} \mathbb{E}[V(\mathbf{z}, x)]$ via search in the semantic space $\mathcal{Z}$.
(ii) Micro-Optimization: Learning a robust instruction-following policy $\pi_{\text{micro}}$ that minimizes the execution divergence given a fixed blueprint $\mathbf{z}^*$.

\subsection{Macro-Micro Architecture}
\begin{figure}[h]
    \centering
    \includegraphics[width=\linewidth]{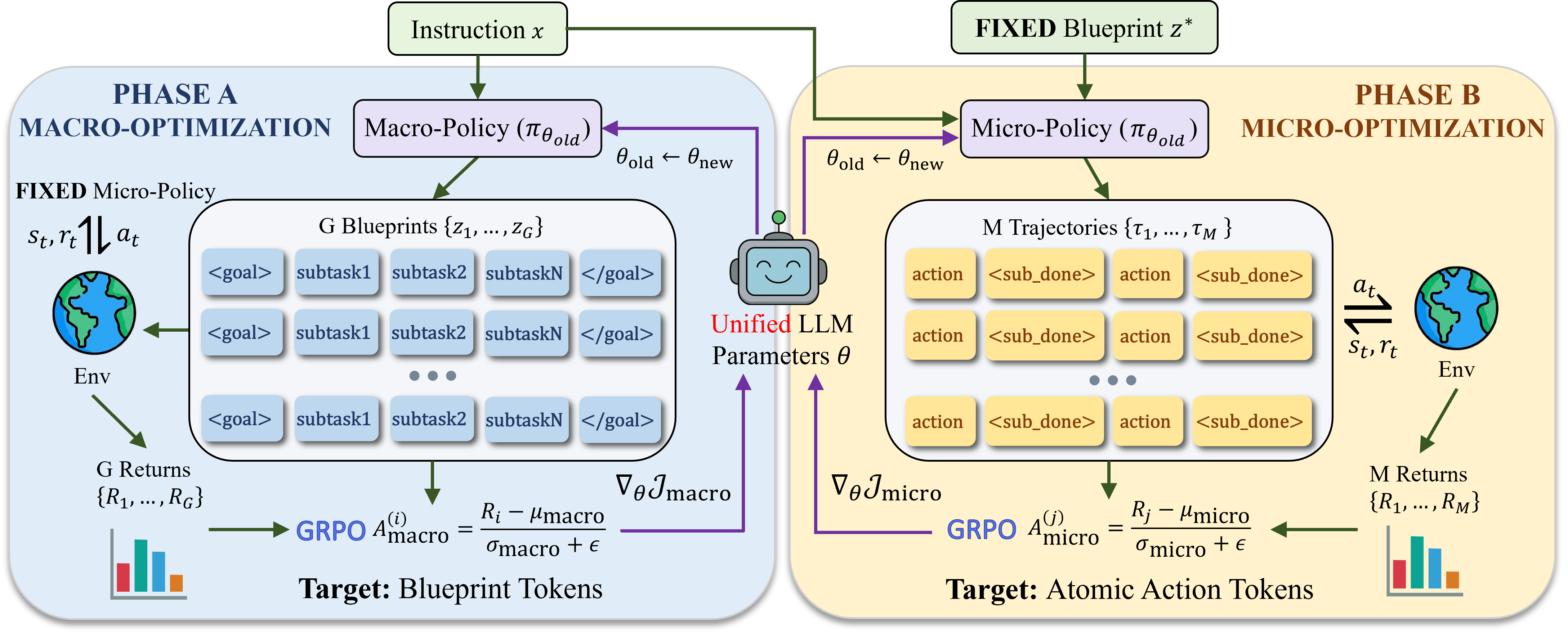}
    \caption{Overall pipeline of HiMAC. Phase A (left) optimizes the macro-policy for blueprint generation using GRPO while freezing the micro-policy. Phase B (right) optimizes the micro-policy for atomic action execution conditioned on a fixed blueprint $z^*$. Both phases update the same underlying LLM parameters $\theta$.}
    \label{fig:algorithm_principle}
\end{figure}
The core premise of HiMAC is decoupling long-horizon reasoning from local execution. First, the Macro-Policy searches over an open-ended natural language space to generate a structured blueprint $z=(z_{1},...,z_{K})$. Each $z_{k}$ represents a free-form semantic milestone, decomposing the global objective into tractable, task-agnostic sub-problems.

Subsequently, the Micro-Policy $\pi_{micro}$ operates as a specialized executor. Conditioned on a selected blueprint $z^{*}$, it sequentially executes the plan. At each timestep $t$, atomic actions are generated based on the currently active sub-goal $z_{k}$:
$a_{t}\sim\pi_{micro}(a_{t}|o_{\le t},z_{k})$.

Sub-goal transitions ($k \leftarrow k+1$) are autonomously triggered when the micro-policy generates a special <sub\_done> termination token. This rigid conditioning acts as a temporal attention mask, explicitly narrowing the agent's context window to the relevant task segment to prevent semantic drift over long horizons. Finally, to handle unexpected environmental stochasticity, we implement a lightweight budget-based fallback: if the executor fails to output <sub\_done> within a step limit $T_{limit}$ or encounters specific failure keywords, the episode is halted and penalized.

\subsection{Hierarchical Policy Optimization}
\label{sec:optimization}
To train the unified policy without auxiliary value networks, we extend GRPO to our hierarchical setting by constructing level-specific comparison groups.
The key insight is that both levels share the same optimization structure: given a group of $N$ sampled outputs with returns $\{R_1, \ldots, R_N\}$, the group-relative advantage for the $i$-th output is computed as:

\begin{equation}
    \hat{A}^{(i)} = \frac{R_i - \mu_{\text{group}}}{\sigma_{\text{group}} + \epsilon}
    \label{eq:advantage}
\end{equation}

where $\mu_{\text{group}}$ and $\sigma_{\text{group}}$ are the mean and standard deviation of returns within the group. The corresponding GRPO-style objective is:

\begin{equation}
    \mathcal{J}(\theta) = \mathbb{E}\left[\frac{1}{N}\sum_{i=1}^{N}\left(\min\left(\rho_i \hat{A}^{(i)},\ \text{clip}(\rho_i, 1{-}\epsilon, 1{+}\epsilon)\hat{A}^{(i)}\right) - \beta\, \mathbb{D}_{\text{KL}}(\pi_\theta \| \pi_{\text{ref}})\right)\right]
    \label{eq:grpo}
\end{equation}

where $\rho_i = \pi_\theta(\cdot) / \pi_{\theta_\text{old}}(\cdot)$ is the importance sampling ratio and $\beta$ controls the KL penalty against the reference model. The two levels instantiate this template differently, as detailed below.

\paragraph{Macro-Objective (Phase A).}
The planner samples a group of $G$ candidate blueprints $\{\mathbf{z}^{(1)}, \ldots, \mathbf{z}^{(G)}\} \sim \pi_{\theta_\text{old}}(\cdot \mid x)$. Each blueprint is evaluated by rolling out the current micro-policy to obtain a trajectory return $R(\mathbf{z}^{(i)})$, which serves as the group signal in Eq.~\eqref{eq:advantage}. Substituting into Eq.~\eqref{eq:grpo} with
\begin{equation*}
    \rho_i^{\text{macro}} = \frac{\pi_\theta(\mathbf{z}^{(i)} \mid x)}{\pi_{\theta_\text{old}}(\mathbf{z}^{(i)} \mid x)}
\end{equation*}
and restricting the gradient to \textbf{blueprint tokens} yields the macro-objective $\mathcal{J}_{\text{macro}}(\theta)$. This setup forces the advantage signal to reflect the \textit{quality of the plan itself}, isolated from the variance introduced by execution stochasticity.

\paragraph{Micro-Objective (Phase B).}
The executor samples a group of $M$ trajectories $\{\tau^{(1)}, \ldots, \tau^{(M)}\} \sim \pi_{\theta_\text{old}}(\cdot \mid x, \mathbf{z}^*)$ conditioned on a fixed high-confidence blueprint $\mathbf{z}^*$. Because all trajectories share the same semantic intent, their return differences are attributable solely to execution quality, forming a valid cohort for group-relative comparison. Substituting into Eq.~\eqref{eq:grpo} with
\begin{equation*}
    \rho_j^{\text{micro}} = \frac{\pi_\theta(\tau^{(j)} \mid x, \mathbf{z}^*)}{\pi_{\theta_\text{old}}(\tau^{(j)} \mid x, \mathbf{z}^*)}
\end{equation*}
and restricting the gradient to \textbf{atomic action tokens} yields the micro-objective $\mathcal{J}_{\text{micro}}(\theta)$.

\subsection{Iterative Co-Evolution Training}
\label{sec:coevolution}

\begin{algorithm}[h]
\caption{HiMAC: Iterative Co-Evolution Training}
\label{alg:himac}
\begin{algorithmic}[1]
\REQUIRE Dataset $\mathcal{D}$, initial policy $\pi_\theta$, 
         reference model $\pi_{\text{ref}}$, 
         group sizes $G$, $M$, iterations $T$
\FOR{iteration $t = 1, \ldots, T$}
    \STATE Sample instruction $x \sim \mathcal{D}$
    \STATE \textbf{// Phase A: Macro-Exploration}
    \STATE Sample blueprint group 
           $\{\mathbf{z}^{(i)}\}_{i=1}^G \sim \pi_{\theta_{\text{old}}}(\cdot \mid x)$
    \FOR{$i = 1, \ldots, G$}
        \STATE Evaluate $R(\mathbf{z}^{(i)})$ via greedy micro-policy 
               rollout \textit{(no grad)}
    \ENDFOR
    \STATE Compute $\{\hat{A}^{(i)}_{\text{macro}}\}$ via Eq.~\eqref{eq:advantage}
    \STATE Update $\theta \leftarrow \theta - \eta \nabla_\theta 
           \mathcal{J}_{\text{macro}}(\theta)$ 
           \textit{(blueprint tokens only)}
    \STATE \textbf{// Phase B: Micro-Adaptation}
    \STATE Select $\mathbf{z}^* = \argmax_i R(\mathbf{z}^{(i)})$
    \STATE Sample trajectory group 
           $\{\tau^{(j)}\}_{j=1}^M \sim \pi_{\theta_{\text{old}}}
           (\cdot \mid x, \mathbf{z}^*)$
    \STATE Compute $\{\hat{A}^{(j)}_{\text{micro}}\}$ via Eq.~\eqref{eq:advantage}
    \STATE Update $\theta \leftarrow \theta - \eta \nabla_\theta 
           \mathcal{J}_{\text{micro}}(\theta)$ 
           \textit{(action tokens only)}
\ENDFOR
\ENSURE Trained hierarchical policy $\pi_\theta$
\end{algorithmic}
\end{algorithm}

Simultaneously optimizing the planner and executor introduces a fundamental non-stationarity: the planner chases a shifting execution capability while the executor adapts to a drifting sub-goal distribution.
To stabilize this bi-level optimization, we introduce an Iterative Co-Evolution strategy (Algorithm~\ref{alg:himac}) that decouples the process into two alternating phases, ensuring each level is updated against a stationary counterpart.

\paragraph{Phase A: Macro-Exploration.}
We optimize $\mathcal{J}_{\text{macro}}(\theta)$ while treating the 
micro-policy as a \textit{fixed} component of the environment. For each 
sampled blueprint $\mathbf{z}^{(i)} \sim \pi_\theta(\cdot \mid x)$, the 
executor evaluates it via greedy decoding in inference mode 
(no gradient tracking), providing a \textit{deterministic} return signal 
$R(\mathbf{z}^{(i)})$ that reflects plan feasibility without execution 
variance.

\paragraph{Phase B: Micro-Adaptation.}
We optimize $\mathcal{J}_{\text{micro}}(\theta)$ conditioned on a high-confidence blueprint:
\begin{equation}
    \mathbf{z}^* = \argmax_{i} \, R(\mathbf{z}^{(i)})
    \label{eq:best_blueprint}
\end{equation}
selected from the group sampled in Phase A. Fixing $\mathbf{z}^*$ as a 
constant condition (via stop-gradient) ensures that return differences 
across the sampled trajectories $\{\tau^{(j)}\}_{j=1}^M$ are attributable 
solely to execution quality, providing a clean learning signal for 
low-level control. Using the highest-return blueprint (rather than a  randomly sampled one) further guarantees the executor trains on coherent, currently achievable plans, avoiding wasted rollouts
(ablated in Sec.~\ref{sec:ablation}).

\section{Experiments}
\label{sec:exp}
In this section, we empirically validate the effectiveness of HiMAC on complex, long-horizon agentic tasks.
Our evaluation focuses on three central hypotheses:
(i) the hierarchical decoupling of planning and execution yields superior performance compared to flat policy architectures;
(ii) the Iterative Co-Evolution strategy effectively stabilizes the non-stationary dynamics inherent in bi-level optimization; (iii) our method achieves these gains with high sample efficiency.

\subsection{Experimental Setup}
\paragraph{Benchmarks and Baselines.}
To ensure a rigorous evaluation aligned with state-of-the-art standards, we conduct experiments on various challenging benchmarks. ALFWorld \cite{shridhar2020alfworld} evaluates embodied decision-making across six distinct task types requiring multi-step reasoning in a simulated household environment.
WebShop \cite{yao2022webshop} tests the agent's ability to navigate a noise-heavy e-commerce website to locate and purchase products matching specific attributes. Sokoban \cite{SchraderSokoban2018} introduces a visually-grounded spatial planning challenge, requiring the agent to execute precise sequential reasoning to navigate and push boxes to designated targets. We compare HiMAC against a comprehensive suite of baselines, including prompt-based methods (ReAct \cite{yao2022react}, Reflexion \cite{shinn2023reflexion}) and recent reinforcement learning advancements (PPO \cite{schulman2017proximal}, RLOO \cite{ahmadian2024back}, GRPO \cite{shao2024deepseekmath} and GiGPO \cite{feng2025group}.

\paragraph{Implementation Details.}
For text-based tasks (ALFWorld and WebShop), we employ Qwen2.5-Instruct \cite{qwen2025qwen25technicalreport} (1.5B and 7B) as the backbone for both the Macro-Policy and Micro-Policy.
For the visually-grounded Sokoban benchmark, we utilize Vision-Language Models (VLMs) including the Qwen2.5-VL \cite{bai2025qwen25vltechnicalreport} (3B and 7B) and Qwen3-VL \cite{bai2025qwen3} (8B) series.
Across all tasks, we maintain a consistent response length of 512 and a rollout group size of $N=8$. For ALFWorld, the maximum prompt length is 2048 with 128 parallel environments; for WebShop, the maximum prompt length is 4096 and episodes are limited to 15 steps; for Sokoban, the maximum prompt length is 1024 with a similar episode limit of 15 steps. The KL-divergence coefficient $\beta$ is fixed at 0.01 across all runs.

\subsection{Main Results}
\begin{table*}[h]
\centering
\caption{Performance on ALFWorld and WebShop. Results are averaged over 3 random seeds. PE refers to prompt engineering. GiGPO$_{\text{w/ std}}$ denotes using $F_{\text{norm}} = \text{std}$, while GiGPO$_{\text{w/o std}}$ uses $F_{\text{norm}} = 1$.}
\resizebox{\textwidth}{!}{
\begin{tabular}{llccccccc|cc}
\toprule
\multirow{2}{*}{\textbf{Type}} & \multirow{2}{*}{\textbf{Method}} & \multicolumn{7}{c}{\textbf{ALFWorld}} & \multicolumn{2}{c}{\textbf{WebShop}} \\
\cmidrule(lr){3-9} \cmidrule(lr){10-11}
& & Pick & Look & Clean & Heat & Cool & Pick2 & All & Score & Succ. \\
\midrule
\multicolumn{11}{c}{\textit{Closed-Source Model}} \\
\midrule
PE & GPT-4o & 75.3 & 60.8 & 31.2 & 56.7 & 21.6 & 49.8 & 48.0 & 31.8 & 23.7 \\
PE & Gemini-2.5-Pro & 92.8 & 63.3 & 62.1 & 69.0 & 26.6 & 58.7 & 60.3 & 42.5 & 35.9 \\
PE & GPT-5 & - & - & - & - & - & - & 65.4 & - & 33.7 \\
PE & Claude Sonnet 4.5 & - & - & - & - & - & - & 82.5 & - & 38.6 \\
\midrule
\multicolumn{11}{c}{\textit{Qwen2.5-1.5B-Instruct}} \\
\midrule
PE & Qwen2.5 & 5.9 & 5.5 & 3.3 & 9.7 & 4.2 & 0.0 & 4.1 & 23.1 & 5.2 \\
PE & ReAct & 17.4 & 20.5 & 15.7 & 6.2 & 7.7 & 2.0 & 12.8 & 40.1 & 11.3 \\
PE & Reflexion & 35.3 & 22.2 & 21.7 & 13.6 & 19.4 & 3.7 & 21.8 & 55.8 & 21.9 \\
RL & PPO & $64.8_{\pm 3.5}$ & $40.5_{\pm 6.9}$ & $57.1_{\pm 4.9}$ & $60.6_{\pm 6.6}$ & $46.4_{\pm 4.0}$ & $47.4_{\pm 1.9}$ & $54.4_{\pm 3.1}$ & $73.8_{\pm 3.0}$ & $51.5_{\pm 2.9}$ \\
RL & RLOO & $88.3_{\pm 3.0}$ & $52.8_{\pm 8.6}$ & $71.0_{\pm 5.9}$ & $62.8_{\pm 8.7}$ & $66.4_{\pm 5.5}$ & $56.9_{\pm 4.7}$ & $69.7_{\pm 2.5}$ & $73.9_{\pm 5.6}$ & $52.1_{\pm 6.7}$ \\
RL & GRPO & $85.3_{\pm 1.5}$ & $53.7_{\pm 8.0}$ & $84.5_{\pm 6.8}$ & $78.2_{\pm 7.9}$ & $59.7_{\pm 5.0}$ & $53.5_{\pm 5.6}$ & $72.8_{\pm 3.6}$ & $75.8_{\pm 3.5}$ & $56.8_{\pm 3.8}$ \\
RL & GiGPO$_{\text{w/ std}}$ & \textbf{94.4$_{\pm 5.9}$} & $67.5_{\pm 4.6}$ & $94.8_{\pm 3.8}$ & $94.4_{\pm 7.8}$ & $79.8_{\pm 4.7}$ & $76.4_{\pm 5.4}$ & $86.7_{\pm 1.7}$ & $83.1_{\pm 1.6}$ & $65.0_{\pm 3.2}$ \\
RL & GiGPO$_{\text{w/o std}}$ & $96.0_{\pm 4.7}$ & $76.5_{\pm 1.9}$ & $91.8_{\pm 5.5}$ & $91.3_{\pm 6.1}$ & $71.7_{\pm 8.4}$ & $79.5_{\pm 7.7}$ & $86.1_{\pm 4.7}$ & $83.5_{\pm 1.8}$ & $67.4_{\pm 4.5}$ \\
\rowcolor{gray!15}
RL& \textbf{HiMAC} &\textbf{98.6$_{\pm 1.5}$}&\textbf{79.9$_{\pm 4.7}$}&\textbf{98.2$_{\pm 1.8}$}&90.5$_{\pm 3.1}$&\textbf{80.1$_{\pm 3.6}$}&\textbf{84.3$_{\pm 5.6}$}&\textbf{89.9$_{\pm 3.4}$}&\textbf{92.2$_{\pm 2.9}$}&\textbf{83.4$_{\pm 1.3}$}\\
\midrule
\multicolumn{11}{c}{\textit{Qwen2.5-7B-Instruct}} \\
\midrule
PE & Qwen2.5 & 33.4 & 21.6 & 19.3 & 6.9 & 2.8 & 3.2 & 14.8 & 26.4 & 7.8 \\
PE & ReAct & 48.5 & 35.4 & 34.3 & 13.2 & 18.2 & 17.6 & 31.2 & 46.2 & 19.5 \\
PE & Reflexion & 62.0 & 41.6 & 44.9 & 30.9 & 36.3 & 23.8 & 42.7 & 58.1 & 28.8 \\
RL & PPO & $92.3_{\pm 4.0}$ & $64.0_{\pm 8.4}$ & $92.5_{\pm 2.4}$ & $89.5_{\pm 7.0}$ & $80.3_{\pm 2.0}$ & $68.8_{\pm 8.3}$ & $80.4_{\pm 2.7}$ & $81.4_{\pm 3.1}$ & $68.7_{\pm 5.1}$ \\
RL & RLOO & $87.6_{\pm 4.3}$ & $78.2_{\pm 8.3}$ & $87.3_{\pm 5.8}$ & $81.3_{\pm 7.6}$ & $71.9_{\pm 5.2}$ & $48.9_{\pm 8.4}$ & $75.5_{\pm 4.6}$ & $80.3_{\pm 3.2}$ & $65.7_{\pm 4.0}$ \\
RL & GRPO & $90.8_{\pm 5.1}$ & $66.1_{\pm 6.7}$ & $89.3_{\pm 5.4}$ & $74.7_{\pm 6.9}$ & $72.5_{\pm 5.4}$ & $64.7_{\pm 7.3}$ & $77.6_{\pm 5.2}$ & $79.3_{\pm 2.8}$ & $66.1_{\pm 3.7}$ \\
RL & GiGPO$_{\text{w/ std}}$ & $97.7_{\pm 1.6}$ & $82.7_{\pm 7.9}$ & $97.8_{\pm 1.6}$ & $83.7_{\pm 7.2}$ & $89.3_{\pm 8.2}$ & $79.2_{\pm 6.6}$ & $90.8_{\pm 1.3}$ & $84.4_{\pm 2.9}$ & $72.8_{\pm 3.2}$ \\
RL & GiGPO$_{\text{w/o std}}$ & $91.8_{\pm 5.4}$ & \textbf{88.6$_{\pm 6.3}$} & $95.9_{\pm 3.2}$ & $90.2_{\pm 2.6}$ & $86.5_{\pm 5.5}$ & $85.2_{\pm 7.5}$ & $90.2_{\pm 2.3}$ & $86.2_{\pm 2.6}$ & $75.2_{\pm 3.8}$ \\
\rowcolor{gray!15}
RL &\textbf{HiMAC} &\textbf{99.1$_{\pm 0.9}$} &88.1$_{\pm 5.9}$&\textbf{98.4$_{\pm 4.5}$}&\textbf{91.9$_{\pm 2.3}$}&\textbf{92.7$_{\pm 4.4}$}&\textbf{89.5$_{\pm 6.1}$}&\textbf{92.1$_{\pm 3.5}$}&\textbf{93.8$_{\pm 2.4}$}&\textbf{84.1$_{\pm 2.9}$}\\
\bottomrule
\end{tabular}
}
\label{tab:performance}
\end{table*}
\paragraph{ALFWorld.}
Table \ref{tab:performance} reports the performance of all methods on ALFWorld and WebShop.
On ALFWorld with the 1.5B backbone, HiMAC achieves an overall success rate of 89.9\%, surpassing the strongest multi-turn RL baseline, GiGPO,
by 3.8\%.
The gains are particularly pronounced on structurally complex task types: Pick2 (multi-target pick, 84.3 vs. 76.4\%) and Clean (98.2 vs. 91.8\%), which require coherent multi-stage planning.
This demonstrates that the Structured Blueprint effectively partitions the long-horizon task into tractable sub-goals,
preventing the context drift that causes flat policies to fail.
Notably, despite having only 1.5B parameters, our HiMAC-trained model achieves an 89.9\% success rate, surpassing the prompt-only performance of the much larger closed-source model Claude Sonnet-4.5 (82.5\%) and demonstrating the efficiency of our hierarchical structure.

Scaling to the 7B backbone, HiMAC achieves an overall success rate of 92.1\%, consistently outperforming all baselines across all six task categories. The improvement over the best flat RL baseline (GiGPO, 90.8\%) reflects that hierarchical decomposition provides complementary benefits beyond those of group-based credit assignment alone.

\paragraph{WebShop.}
The WebShop benchmark presents a substantially more challenging setting due to its large action space and noisy observation signals. HiMAC demonstrates exceptional gains here: with the 1.5B backbone, it achieves a success rate of 83.4\% and a score of 92.2\%, surpassing GiGPO by a remarkable \textbf{16.0\%} in success rate (67.4\%).
We attribute this large margin to the Structured Blueprint's ability to decompose the web navigation task into semantic milestones, preventing the executor from losing track of the overall purchasing intent across long interaction sequences.
With the 7B backbone, HiMAC achieves a success rate of 84.1\% and a score of 93.8\%, again outperforming all RL baselines.

\paragraph{Sokoban.}
As shown in Table \ref{tab:sokoban}, on the Sokoban benchmark using the Qwen2.5-VL-7B backbone, HiMAC achieves a success rate of 87.5\% and a score of 6.70, outperforming GiGPO (82.8\%, 5.27) by 4.7 points in success and 1.43 points in score.
This result demonstrates that the hierarchical planning paradigm generalizes beyond text-based agentic tasks to visually-grounded puzzle environments, where the Macro-Policy effectively learns to decompose box-pushing sequences into structured subgoal chains.

\paragraph{Prompt-Based vs. RL-Based Methods.}
Across all benchmarks, prompt-based methods (ReAct, Reflexion) lag significantly behind RL-based approaches, confirming that pre-trained LLMs require parameter-level adaptation to master long-horizon sequential tasks. Among RL methods, the overall performance comparison suggests that finer-grained credit assignment and hierarchical structure are both independently beneficial, and their combination in HiMAC yields the strongest performance.

\begin{wraptable}{r}{0.6\textwidth}
    \vspace{-15pt}
  \centering
  \caption{Performance on the Sokoban benchmark. 'Succ.' and 'Score' refers to success rate and test score respectively.}
  \label{tab:sokoban}
  \setlength{\tabcolsep}{3pt}
  \resizebox{\linewidth}{!}{
  \begin{tabular}{l c c c c c c}
    \toprule
    \multirow{2}{*}{\textbf{Method}} & \multicolumn{2}{c}{\textbf{Qwen2.5-VL-3B}} & \multicolumn{2}{c}{\textbf{Qwen2.5-VL-7B}} & \multicolumn{2}{c}{\textbf{Qwen3-VL-8B}} \\
    \cmidrule(lr){2-3} \cmidrule(lr){4-5} \cmidrule(lr){6-7}
    & \textbf{Succ.} & \textbf{Score} & \textbf{Succ.} & \textbf{Score} & \textbf{Succ.} & \textbf{Score} \\
    \midrule
    Prompt             & 14.1 & -0.88 & 25.0 & -0.26 & 21.9 & -0.95 \\
    GRPO            & 76.2 & 4.59 & 83.9 & 5.39 & 82.9 & 5.24 \\
    GiGPO           & 81.1 & 4.86 & 82.8 & 5.27 & 78.1 & 4.67 \\
    \textbf{HiMAC}  & \textbf{83.8} & \textbf{5.31} & \textbf{87.5} & \textbf{6.70} & \textbf{90.6} & \textbf{7.32} \\
    \bottomrule
  \end{tabular}
  }
  \vspace{-15pt}
\end{wraptable}

\subsection{Ablation Studies}
\label{sec:ablation}

To isolate the contribution of each core design decision in HiMAC, we conduct a systematic ablation study on ALFWorld and WebShop using the Qwen2.5-7B-Instruct backbone. We evaluate four variants, each removing or replacing a single component of the full system. Results are reported in Table \ref{tab:ablation1}.

\paragraph{w/o Hierarchy.}
In this variant, we remove the entire hierarchical structure—both the structured blueprint generation and the level-specific group construction—and instead apply a single GRPO objective over the complete, flat trajectory.
Under this flat formulation, blueprint-level credit and execution-level credit are conflated into a single advantage estimate, making it impossible to attribute outcome variance to planning decisions versus execution decisions independently. Note that this variant also implicitly removes the Iterative Co-Evolution strategy, since the alternating phase structure presupposes the existence of a two-level hierarchy.
The performance drop of 14.5\% on ALFWorld and 18.0\% on WebShop confirms that the two-level group construction provides strictly more precise credit assignment than a flat rollout group. Without this separation, the Macro-Policy receives a noisy learning signal contaminated by execution stochasticity, slowing convergence and degrading plan quality.
\begin{table}[h]
    \centering
    \vspace{-15pt}
    \caption{Ablation study on ALFWorld and Webshop (Qwen2.5-7B-Instruct).}
    \begin{tabular}{c|ccc}
    \toprule
        Variant & ALFWorld (\%)&WebShop Score& WebShop Succ.(\%) \\
        \midrule
        HiMAC (Full) & 92.1 &93.8&84.1\\
        w/o hierarchy (Flat GRPO) & 77.6&79.3&66.1\\
        w/o Iterative Co-Evolution &85.3&86.7&74.8\\
        w/o <sub\_done>&88.2&90.1&79.8 \\
        Random Blueprint&89.7&91.6&81.5 \\
        \bottomrule
    \end{tabular}
    \label{tab:ablation1}
    \vspace{-15pt}
\end{table}

\paragraph{w/o Iterative Co-Evolution.}
Here we abandon the Iterative Co-Evolution strategy and instead update both the Macro-Policy and Micro-Policy simultaneously within each training step.
As discussed in Section \ref{sec:coevolution}, this introduces non-stationary dynamics: the planner optimizes sub-goals for an executor whose capabilities are shifting within the same gradient step, and vice versa. The resulting 6.8\% degradation on ALFWorld validates our alternating optimization motivation—by freezing one level while updating the other, HiMAC converts the unstable bi-level problem into a sequence of stationary single-level problems. The larger relative drop on WebShop (9.3\%) suggests that non-stationarity is particularly harmful in environments with longer horizons and noisier reward signals, where gradient interference accumulates more severely.

\paragraph{w/o <sub\_done> token.}
Rather than allowing the Micro-Policy to autonomously signal sub-goal completion via the <sub\_done> termination token, this variant switches to a fixed step budget: each sub-goal is allocated a pre-determined number of steps T\_{\text{fixed}} = T\_{\text{limit}},
after which the system advances to the next sub-goal regardless of execution state. This eliminates the model's ability to self-regulate pacing based on task difficulty.
The resulting 3.9\%/4.3\% drop on ALFWorld/Webshop reveals that adaptive termination is a critical capability: some sub-goals (e.g., locating an object in a cluttered environment) require variable step counts that a fixed budget cannot accommodate. When the budget is too short, the executor is interrupted mid-task; when too long, it wastes steps after completion and risks drifting into erroneous actions. The learnable
<sub\_done> mechanism enables the Micro-Policy to develop a reliable sense of task completion, contributing meaningfully to the overall system's robustness.

\begin{table}[h]
\centering
\caption{Comprehensive sample efficiency analysis across benchmarks (7B backbone). We report both the convergence speed (approximate iterations required to reach a target success threshold) and the final converged success rate.}
\label{tab:sample_efficiency}
\begin{tabular}{lcccccc}
\toprule
\multirow{2}{*}{\textbf{Benchmark (Target)}} & \multicolumn{3}{c}{\textbf{Iters to Target} $\downarrow$} & \multicolumn{3}{c}{\textbf{Final Succ. (\%)} $\uparrow$} \\
\cmidrule(lr){2-4} \cmidrule(lr){5-7}
 & \textbf{HiMAC} & GiGPO & GRPO & \textbf{HiMAC} & GiGPO & GRPO \\
\midrule
\textbf{ALFWorld} (75\%) & \textbf{$\sim$110} & $\sim$115 & $\sim$150 & \textbf{92.1} & 90.2 & 77.6 \\
\textbf{WebShop} (65\%)  & \textbf{$\sim$220} & $\sim$230 & $\sim$380 & \textbf{84.1} & 72.8 & 65.7 \\
\textbf{Sokoban} (80\%)  & \textbf{$\sim$180} & $\sim$195 & $\sim$210 & \textbf{87.5} & 82.8 & 83.9 \\
\bottomrule
\end{tabular}
\vspace{-15pt}
\end{table}

\paragraph{Random Blueprint.}
In the full HiMAC system,
Phase B conditions the Micro-Policy on a high-confidence blueprint
$z^*=argmax_i R(z^{(i)})$, i.e., the blueprint from the sampled group that achieved the highest trajectory return.
This design ensures that the executor is trained on plans that are both coherent and practically achievable under the current planner's capability.
In this variant, we replace this selection with a randomly sampled blueprint from the same group,
regardless of its return.
The performance drops by 2.4\% on ALFWorld and 2.6\% on WebShop, demonstrating that the quality of the conditioning blueprint during Micro-training has a direct and significant impact on execution learning.
When the executor is trained on low-quality or incoherent blueprints, it acquires execution patterns conditioned on unreliable semantic intents, introducing noise into the Micro-Policy's gradient signal and slowing convergence. 
The high-confidence selection thus serves as an implicit quality filter, ensuring that the executor's "bootcamp" is grounded in feasible plans and avoiding the waste of rollout budget on unexecutable decompositions.

\subsection{Analysis}

\paragraph{Sample Efficiency.}
Table~\ref{tab:sample_efficiency} compares the sample efficiency of HiMAC against existing methods by measuring performance achieved under limited training iterations. HiMAC consistently reaches strong performance levels with fewer training samples, demonstrating faster convergence than flat-policy baselines.

This improvement mainly stems from the hierarchical design. By decomposing decision-making into blueprint generation and execution, HiMAC reduces the effective exploration space and enables more stable credit assignment through hierarchical group-relative advantages. In addition, the Iterative Co-Evolution strategy introduces a natural learning curriculum, allowing the agent to progressively learn executable plans instead of wasting samples on infeasible trajectories.

These results indicate that HiMAC not only improves final performance but also utilizes interaction samples more efficiently, which is particularly important for long-horizon agent training.

\paragraph{Qualitative Analysis of Learned Blueprints.}
\begin{wrapfigure}{r}{0.7\textwidth}
    \centering
    \vspace{-20pt}
    \includegraphics[width=\linewidth]{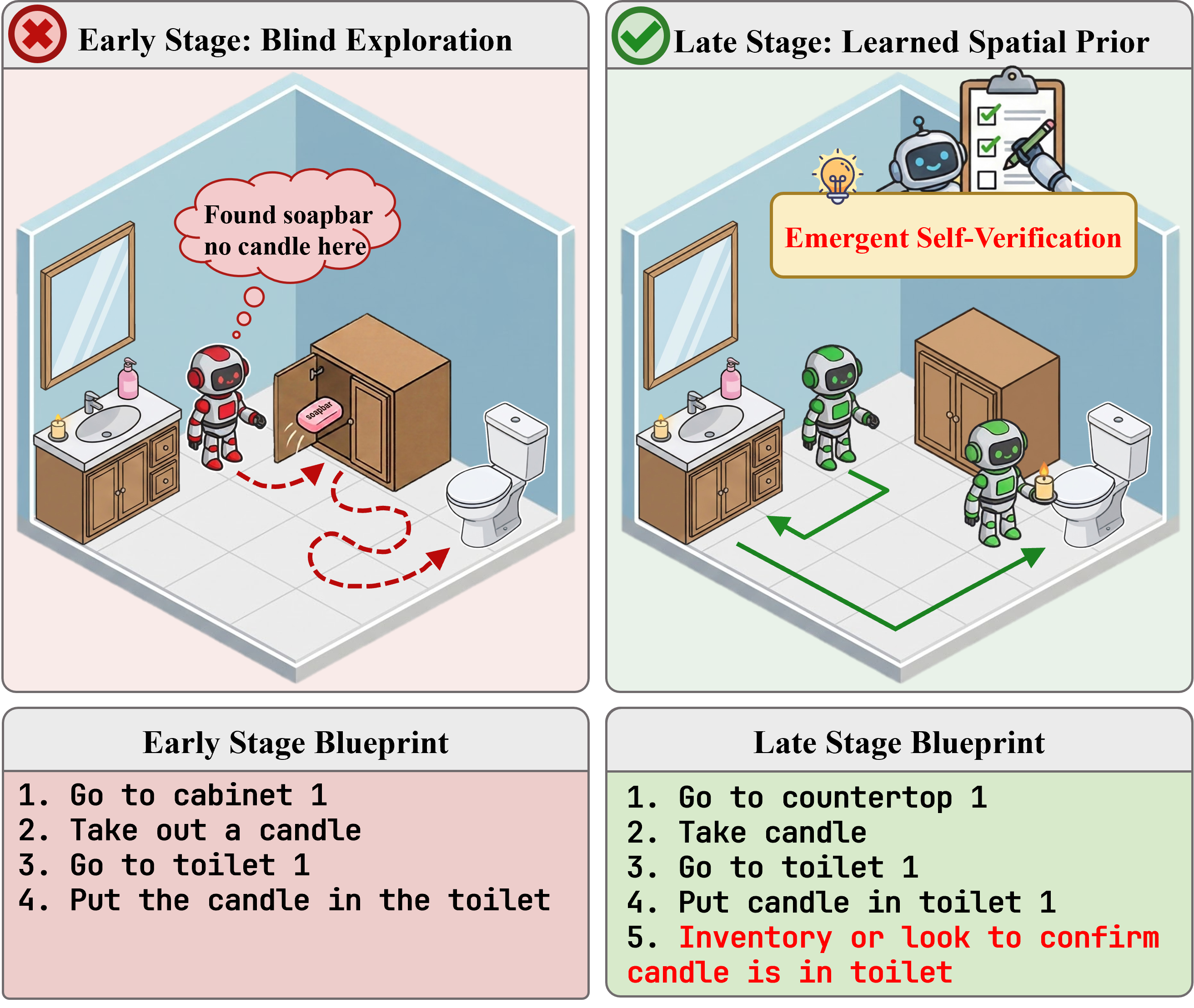}
    \caption{Evolution of HiMAC's planning and emergent behaviors. For the task "put a candle in toilet": (Left) Early training shows blind exploration due to lacking spatial priors. (Right) Post-convergence, the planner efficiently navigates to the target and develops emergent self-verification.}
    \label{fig:blueprint}
    \vspace{-15pt}
\end{wrapfigure}
To gain interpretable insight into HiMAC's planning behavior, Figure \ref{fig:blueprint} visualizes representative blueprints generated by the Macro-Policy at early (iteration 19) and late (iteration 239) training stages on the ALFWorld task "put a candle in the toilet". Early in training, blueprints reflect a lack of spatial common sense and rely on naive, open-loop exploration (e.g., wastefully searching a closed cabinet for a candle). After convergence, the generated blueprints not only exhibit accurate spatial priors by directly locating the target on the countertop, but also demonstrate a striking emergent self-verification mechanism. Specifically, the mature planner autonomously appends an observation step (e.g., "Inventory or look to confirm") to validate task success. This critical transition—from blind execution to closed-loop, structurally complex planning—clearly visualizes the emergent co-evolution of planner complexity and executor proficiency. This qualitative evidence strongly supports our co-evolutionary curriculum hypothesis.

\paragraph{Scalability with Model Size.} Comparing the 1.5B and 7B variants of HiMAC reveals a consistent 2–3 percentage point improvement in overall ALFWorld success rate (89.9\% → 92.1\%) and improvement in WebShop (83.4\% → 84.1\%), suggesting that HiMAC's hierarchical framework scales gracefully with backbone capacity and that the architectural gains are largely orthogonal to model scale. This property is practically significant, as it implies that HiMAC can deliver strong performance even in resource-constrained deployment scenarios.

\paragraph{Emergent Curriculum.}
By alternating these two phases every iteration, HiMAC fosters an 
emergent curriculum without any explicit difficulty scheduling. Early 
in training, only simple blueprints yield positive returns under the 
weak executor, so the group-relative advantage in Eq.~\eqref{eq:advantage} 
naturally pushes the planner toward short-horizon decompositions. As 
executor proficiency improves, more complex blueprints begin to yield 
higher returns, incentivizing the planner to progressively explore 
richer strategies.
This co-evolutionary dynamic enables the stable emergence of hierarchical cognitive structures across the full training horizon.

\section{Conclusion}
\label{sec:conclu}

In this paper, we presented HiMAC, a novel hierarchical reinforcement learning framework designed to enable Large Language Models to robustly handle long-horizon agentic tasks. By formulating the reasoning process as a bi-level optimization problem, HiMAC explicitly decouples strategic planning from tactical execution, effectively mitigating the exponential error accumulation inherent in traditional flat policies.
To address the optimization challenges of such decoupled architectures, we introduced two key technical innovations: a Critic-Free Hierarchical Group-Based Optimization objective and an Iterative Co-Evolution training strategy. These mechanisms allow the agent to learn precise credit assignment without unstable value networks and ensure the synchronized improvement of both the planner and the executor.
Extensive experiments on complex benchmarks, including ALFWorld, WebShop and Sokoban, demonstrate that HiMAC achieves state-of-the-art success rates with superior sample efficiency compared to strong baselines. Our findings suggest that imposing structural inductive biases—specifically the Macro-Micro separation—is crucial for scaling long-horizon agents towards real-world complexity. Future work will explore applying HiMAC to more open-ended environments and investigating the transferability of learned blueprints across different domains.

\clearpage  


%
%
\bibliographystyle{splncs04}
\bibliography{main}
\end{document}